\PassOptionsToPackage{numbers,sort&compress}{natbib}
\documentclass[journal,print]{IEEEtran}
\usepackage{generic}
\usepackage{cite}
\usepackage{amsmath,amssymb,amsfonts}
\usepackage{graphicx}
\usepackage{algorithm,algorithmic}
\usepackage{color}
\usepackage{placeins}
\usepackage{subcaption}
\usepackage{orcidlink}
\restylefloat{figure}
\restylefloat{table}
\usepackage{xcolor}
\usepackage{hyperref}
\makeatletter
\let\NAT@parse\undefined

\def\@citex[#1]#2{%
  \let\@citea\@empty
  \@cite{\@for\@citeb:=#2\do
    {\@citea\def\@citea{], [}%
     \edef\@citeb{\expandafter\@firstofone\@citeb\@empty}%
     \if@filesw\immediate\write\@auxout{\string\citation{\@citeb}}\fi
     \@ifundefined{b@\@citeb}{\mbox{\reset@font\bfseries ?}%
       \G@refundefinedtrue
       \@latex@warning
         {Citation `\@citeb' on page \thepage \space undefined}}%
       {\hbox{\csname b@\@citeb\endcsname}}}}{#1}}

\renewcommand\@cite[2]{{\color{blue}[{#1\if@tempswa , #2\fi}]}}
\makeatother

\hypersetup{
  colorlinks=true,
  citecolor=blue,
  linkcolor=black,
  urlcolor=blue
}
\usepackage{booktabs}
\usepackage{float}
\usepackage{tabularx}
\usepackage{array}
\usepackage{textcomp}
\usepackage{multirow}
\usepackage{adjustbox}
\def\BibTeX{{\rm B\kern-.05em{\sc i\kern-.025em b}\kern-.08em
    T\kern-.1667em\lower.7ex\hbox{E}\kern-.125emX}}

\begin{document}
\title{Understanding Key Features of Time Series Foundation Models from Epidemic Forecasting}

\author{Alireza~Jafari,
        Judy~Fox,
        Geoffrey~C.~Fox,
        Madhav~Marathe,
        and Aniruddha~Adiga
\thanks{This manuscript is a preprint version of the work. }
\thanks{Alireza Jafari is with the Department of Computer Science, School of Engineering and Applied Science, University of Virginia, Charlottesville, VA 22904 USA (e-mail: jrp5td@virginia.edu).}
\thanks{Judy Fox is with the School of Data Science, University of Virginia, Charlottesville, VA 22904 USA, and also with the Department of Computer Science, School of Engineering and Applied Science, University of Virginia, Charlottesville, VA 22904 USA (e-mail: ckw9mp@virginia.edu).}
\thanks{Geoffrey C. Fox is with the Biocomplexity Institute and the Department of Computer Science, School of Engineering and Applied Science, University of Virginia, Charlottesville, VA 22904 USA (e-mail: vxj6mb@virginia.edu).}
\thanks{Madhav Marathe is with the Biocomplexity Institute and the Department of Computer Science, School of Engineering and Applied Science, University of Virginia, Charlottesville, VA 22904 USA, and also with the Department of Electrical and Computer Engineering, School of Engineering and Applied Science, University of Virginia, Charlottesville, VA 22904 USA, by courtesy (e-mail: mvm7hz@virginia.edu).}
\thanks{Aniruddha Adiga is with the Biocomplexity Institute, University of Virginia, Charlottesville, VA 22904 USA (e-mail: aniruddha@virginia.edu).}
}

\maketitle

\begin{abstract}
Seasonal influenza infects millions of people and causes substantial morbidity and mortality in the United States each year, making accurate short-term forecasting a core public-health need. Reliable forecasts of epidemic time series can inform vaccination timing, hospital staffing, and resource allocation, yet the comparative behavior of modern forecasting architectures on infectious-disease surveillance data remains insufficiently characterized. 
We address this gap through a systematic evaluation of regional influenza forecasting using influenza-like illness surveillance and influenza-associated hospitalization time series under both temporal and spatial generalization settings for 1–4-week-ahead prediction. We compare classical neural network architectures, numerical transformer-based models, pretrained time series foundation models, and LLM-based forecasting approaches. Across tasks, we demonstrate that a mixture-of-experts model that fuses multiple pretrained forecasters achieves the strongest overall performance, indicating that heterogeneous pretrained representations provide complementary predictive information. Our results further show that numerical transformer-based models produce reliable forecasts, while pretraining provides the largest gains at longer horizons, particularly when the pretraining domain is mechanistically aligned with influenza dynamics. In contrast, LLM-based time series methods underperform relative to numerical forecasters in this setting. Finally, we examine hospitalization information as both an auxiliary covariate and a pretraining source. Hospitalization signals provide complementary improvements in selected settings and clarify when additional surveillance streams enhance the robustness of multi-horizon forecasting. These findings provide actionable guidance on model selection, pretraining strategy, and auxiliary-signal use for influenza surveillance and preparedness.

\end{abstract}

\vspace{5mm}
\begin{IEEEkeywords}
Epidemic Forecasting, Foundation Model, Influenza-like illness, Mixture of Experts, Multi-horizon Prediction, Spatiotemporal Generalization, Time Series Forecasting, Transformers
\end{IEEEkeywords}

\vspace{5mm}
\section{Introduction}
\label{sec:introduction}
Forecasting the spatiotemporal spread of seasonal influenza is a long–standing challenge with direct repercussions for public health. In the United States alone, influenza infects tens of millions of people and contributes to tens of thousands of deaths each year~\cite{reed2015estimating,reich2019collaborative}. Accurate region-specific, short-term forecasts of influenza-like illness (ILI) and influenza-associated hospitalizations help public-health agencies time vaccination campaigns, hospitals plan staffing and bed capacity, and supply chains allocate antivirals and personal protective equipment.

Over the past decade, the CDC FluSight and COVID-19 Forecast Hub initiatives have crystallized these needs into concrete forecasting tasks, bringing together academic, governmental, and industrial teams in a common framework~\cite{cramer2022covidhubdataset}. These hubs emphasize multi-horizon forecasting, while also curating diverse epidemic datasets—supporting tasks such as pretraining, ILI prediction, and hospitalization forecasting—that together have become de facto benchmarks for evaluating epidemic models. At the same time, the methodological landscape has diversified: traditional mechanistic and statistical models remain widely used, while deep learning approaches—including time series foundation models (FM)—have shown strong performance on generic benchmarks spanning finance, earthquake, hydrology, and other domains~\cite{Junyang2026, ansari2024chronos, jafari2022gcnet, jafari2024time, jafari2022netpred}.

Despite this progress, the role of modern time series foundation models in operational epidemic forecasting is still unclear. Recent works fine-tune transformers or time series foundation models on COVID-19 or ILI data and report promising results~\cite{shome2021covid, islam2023interpreting, wu2022timesnet, ansari2024chronos, jin2024timellmtimeseriesforecasting}, but existing evaluations share several limitations. First, they often rely on a single, country-level series with limited history, preventing models from learning the multi-season regional patterns that drive real-world planning. Second, spatial generalization is typically untested: models are trained and evaluated on the same locations, so it is unknown whether a backbone learned on one region can transfer to others with different demographics, healthcare-seeking behavior, or reporting practices. Third, experimental designs frequently use ad-hoc horizons (e.g., 24–60 weeks ahead) and ignore data revisions, diverging from CDC practice, where operational targets are 1–4 weeks ahead and reporting backfill is substantial~\cite{mcgowan2019collaborative}. Finally, different studies adopt heterogeneous data pipelines and report inconsistent rankings for key baselines, making it difficult to disentangle genuine modeling advances from differences in preprocessing or tuning.

% In parallel, emerging time series foundation models themselves are not designed with epidemiology in mind. Architectures such as Chronos and TimeLLM treat time series as sequences of tokens processed by large language-style backbones~\cite{ansari2024chronos, jin2024timellmtimeseriesforecasting, jin2024timellmtimeseriesforecasting}. While powerful on broad, multi-domain benchmarks, they typically assume fixed-shape univariate or homogeneous multivariate inputs and do not natively accommodate rich epidemiological covariates such as ILI, hospitalizations, and regional features. Other foundation-style architectures such as PatchTST, iTransformer, and TimesNet are strong numeric baselines, but prior epidemic case studies have evaluated them only in stylized settings (single series, long horizons) without comparing them directly to LLM-based models or to hybrid ensembles.

In this paper, we build a region-level influenza point-forecasting study to compare the core forecasting behavior of modern time-series models under realistic public-health constraints. We compile and standardize weekly Influenza-Like Illness (ILI) and influenza-associated hospitalization time series at the U.S. HHS-region level, and we evaluate models on the operational 1–4-week horizon used in FluSight-style forecasting. Our evaluation is deliberately two-dimensional: temporal (within-region) generalization tests forward-in-time forecasting for the same regions, while spatial (across-region) generalization holds out entire regions to quantify geographic transfer and distribution shift. Using a consistent preprocessing and training pipeline and reporting both MSE and NNSE, we systematically compare 17 deep forecasting models spanning large task-specific models and modern foundation-style models as well as LLM-based time-series foundation models.

Our main contributions are:

\begin{itemize}
    \item We evaluate time-series foundation models for both influenza-like illness (ILI) and influenza-associated hospitalization forecasting under two explicit generalization regimes, temporal and spatial evaluations using 1–4-week multi-horizon targets. This directly tests whether pretrained temporal priors transfer to operational epidemic forecasting, and quantifies the cost of across-region distribution shift.

    \item We explore a broad range of machine learning architectures and show that mixture-of-experts–style is the most effective strategy for this problem in our study. In particular, our previously introduced MultiFoundationCore model achieves the best performance across our ILI experiments, underscoring the value of expert fusion for robust multi-horizon forecasting. We also provide a controlled comparison across model families (classical neural forecasters, numeric time-series Transformers/foundation models, and LLM-style time-series methods). 

    \item We analyze the impact of pretraining for epidemic forecasting across diverse sources, including mechanistically related signals (e.g., influenza hospitalizations) and large generic time-series corpora (e.g., M4 and TrafficL). These ablations provide practical guidance on when pretraining is most beneficial: improvements are largest at longer lead times (weeks 3–4) and are strongest when the pretraining domain is mechanistically aligned with the target task.

    \item We study the effects of retraining frequency, length of historical data, and the use of mechanistically linked auxiliary signals. We show how related variables (ILI $\leftrightarrow$ hospitalizations) improve forecasts when used as a pretraining domain and/or as an auxiliary input stream, clarifying which integration mode provides the largest gains.
    
    \item We release the code and cleaned data, model configurations, training scripts, and pretrained checkpoints under an open license, providing a reusable testbed for future work on epidemic forecasting with time series foundation and hybrid models. \footnote{
\href{https://github.com/alireza-jafari/Epidemic-Times-Series-Foundation-Models-Benchmark}
{https://github.com/alireza-jafari/
Epidemic-Times-Series-Foundation-\\ Models-Benchmark}
}
\end{itemize}

\section{Related Work}
\label{sec:related-work}

We explore related work on time series epidemic forecasting along three dimensions: (i) mechanistic and statistical models that explicitly encode disease dynamics, (ii) deep learning models for time series forecasting, and (iii) time series foundation models ~\cite{Survey2022, disease_forecasting2022, causalGNN2022, zheng2023enhancing, pathogens11020185, BMC_Public2019}.

Traditional mechanistic approaches, such as SIR and SEIR models, simulate the spread of disease by partitioning the population into compartments based on disease states (e.g., susceptible, exposed, infected, recovered). These models have been widely used to study the dynamics of epidemics and assess the impact of interventions~\cite{SEIR2020, PNAS2024, WSC2021, YANG2021100501}. They are particularly valued for their interpretability and ability to incorporate domain knowledge, such as transmission rates or seasonality. However, they often struggle with scalability and flexibility in rapidly changing real-world settings. Another class of traditional models is statistical methods, which include autoregressive models (AR, ARIMA), generalized linear models (GLMs), and hierarchical Bayesian models. They offer interpretable structure and are often used when data are sparse or highly aggregated. Many were standard in early FluSight systems and remain strong baselines~\cite{math11143069, 2022SR}.

Furthermore, some hybrid models integrate disease-related modules into deep-learning frameworks, especially Graph Neural Networks~(GNNs) and Recurrent Neural Networks~(RNNs), to specify the spatiotemporal epidemic context~\cite{disease_forecasting2022, causalGNN2022, Aniruddha2023, coalgnn20, xie2022epignn, epicola23, RESEAT2023, SAIFluNet}. Many of these works integrate SEIR-style components with GNNs and RNNs to construct disease forecasting models, and have been shown to outperform traditional baselines on pandemic disease hospitalization forecasting. In ColaGNN~\cite{coalgnn20}, the authors design a cross-location attention mechanism to learn time series embeddings for long-term ILI prediction and to update graphs dynamically. 
% Epi-ColaGNN~\cite{epicola23} further integrates a fundamental epidemic component, the next-generation matrix (NGM), into the ColaGNN architecture. In EpiGNN~\cite{xie2022epignn}, a Region-Aware Graph Learner uses transmission risk, geographical dependencies, and temporal information to better capture spatiotemporal dependencies.

Recent advances in deep learning for time series forecasting provide the main building blocks for our epidemic forecasting framework \cite{ansari2024chronos, bai2018empiricalevaluationgenericconvolutional, das2023long, wu2022timesnet, nie2023time, jin2024timellmtimeseriesforecasting, liu2023itransformer}. A line of related work considers non-transformer deep architectures. Temporal Fusion Transformers (TFT)~\cite{lim2021temporal} combine LSTM encoders with static covariate encoders, variable-selection networks, and interpretable attention to produce multi-horizon probabilistic forecasts, originally evaluated on electricity, retail, and traffic datasets.
TiDE~\cite{das2023long} is a purely MLP-based encoder–decoder that jointly ingests past targets and covariates, offering competitive accuracy with transformers while being substantially faster and enjoying theoretical guarantees in linear dynamical systems.
Temporal Convolutional Networks (TCNs)~\cite{bai2018empiricalevaluationgenericconvolutional} use stacks of causal, dilated 1D convolutions to capture multi-scale temporal dependencies; while originally proposed for action segmentation, TCN-style architectures have become standard baselines for time series forecasting.
In principle, TFT and TiDE are well-suited to epidemic forecasting because they natively handle exogenous covariates and multi-step horizons. In practice, however, when these models appear as baselines in the long-term forecasting literature, they are almost always trained without epidemiologically relevant covariates.

A very recent line of work studies pretrained or “foundation” models that aim to reuse knowledge across domains. 
Chronos~\cite{ansari2024chronos} treats time series as a language: it scales and quantizes values into a discrete vocabulary, then trains T5-style transformers with a cross-entropy loss on large multi-domain corpora, achieving strong zero-shot probabilistic forecasts on unseen series.
TimesNet~\cite{wu2022timesnet}, introduced as a general time series architecture, reshapes 1D sequences into multiple period-based 2D tensors and applies convolutional blocks to model local temporal “patches” and their interactions, achieving strong results across a wide range of forecasting and classification tasks and introducing the now-standard ILI benchmark used in many later papers.
TimesNet also popularizes a specific ILI benchmark, which includes a weekly time series of ILI at the country level. For this dataset they fix the input length to 36 weeks and use prediction horizons $T \in \{24, 36, 48, 60\}$. While this configuration is convenient for benchmarking, from an epidemiological perspective, this design has two shortcomings: (i) the ILI data are aggregated at the country level and lack the regional heterogeneity crucial for public-health decision making, and (ii) forecasting 24–60 weeks ahead leaves relatively few independent test periods and does not match CDC practice, where operational targets are current-week to 3-week-ahead incidence or hospitalizations~\cite{reich2019collaborative, CDCFluSight2023}. 
 
Several subsequent deep learning models adopt this exact ILI setup. PatchTST~\cite{nie2023time} builds on the TimesNet benchmark and proposes a patch-based Transformer: it segments each univariate series into patches, treats them as tokens, and shares a channel-independent Transformer across variables, achieving state-of-the-art performance across ETT, Electricity, Weather, Traffic, and the same ILI configuration. iTransformer~\cite{liu2023itransformer} inverts the usual time/feature roles, treating each variate as a token and applying self-attention across variables to better capture multivariate correlations; when evaluated on ILI, it again follows the TimesNet protocol (single national series, 36-step lookback, 24–60-week horizons). Time-LLM~\cite{jin2024timellmtimeseriesforecasting} goes further by reprogramming frozen large language models: numeric windows are mapped into “text prototypes,” prompts are prepended, and the LLM output is projected back to numerical forecasts. Time-LLM reports state-of-the-art average performance on the standard long-term TimesNet benchmark suite, including the ILI dataset, and claims clear gains over strong numerical baselines such as TimesNet and PatchTST under this protocol.

Despite their promise, language-style time-series foundation models have important limitations for ILI forecasting. A central concern is inherited architectural mismatch: models developed for discrete symbolic sequences may not preserve the continuous numerical structure of epidemic surveillance data, where amplitude, slope, phase, local acceleration, seasonality, reporting backfill, regional heterogeneity, and disease progression are all directly meaningful. Quantization, prompt reprogramming, or mapping numerical windows into language-like token spaces can therefore weaken the inductive biases needed for short-horizon epidemic forecasting. At the same time, many reported gains rely on a narrow and unrealistic experimental design: a single short country-level ILI series, no spatial structure or revision effects, and very long horizons of 24–60 weeks that diverge from CDC-style operational forecasting practice. Moreover, different papers report inconsistent baseline numbers and rankings for models such as TimesNet and PatchTST, even when claiming similar settings, making it difficult to separate genuine modeling advances from implementation or tuning artifacts. Thus, claims that reprogrammed LLMs or general foundation models decisively outperform specialized numerical time-series models on ILI remain insufficiently supported under realistic short-horizon, multi-region epidemic forecasting settings.

Furthermore, most current time series foundation models—including Chronos and Time-LLM—are architected around fixed-shape sequences of the target variable and do not natively support rich exogenous or spatial covariates. Their input layers and tokenization schemes are designed for univariate or homogeneous multivariate targets, so incorporating additional signals (hospitalizations, demographics, vaccination coverage, regional features, etc.) typically requires ad-hoc adapters or side models and is not supported in their default evaluation pipelines. As a result, important epidemiological information is often ignored or only partially used, which is a critical limitation for ILI and hospitalization forecasting.

To address these limitations, we previously introduced MultiFoundationCore (MFC) as a time series foundation model built around a mixture-of-experts design that can integrate multiple heterogeneous forecasters within a unified, reusable framework~\cite{jafari2024time, Junyang2026}. In MFC, strong pretrained numeric backbones (and, optionally, complementary task-specific pattern models) are treated as experts, and a learned fusion module combines their representations to produce stable multi-horizon forecasts. Importantly, MFC is designed to support auxiliary signals via explicit input streams (and, in some variants, additional spatial/exogenous modules), mitigating a key limitation of many off-the-shelf time series foundation models that cannot natively exploit epidemiological covariates without ad-hoc adapters. Our prior work explored multiple MFC variants that differ in the expert set, the fusion mechanism, and the use of auxiliary inputs~\cite{jafari2024time}; in the present study, we operationalize these ideas for influenza forecasting and analyze when MoE-style fusion and auxiliary signals provide the largest gains.

\section{Data}
\label{sec:data}

\begin{figure*}[h]
  \centering
  % Top row: two side-by-side
  \begin{subfigure}[t]{0.5\textwidth}
    \centering
    \includegraphics[width=\linewidth]{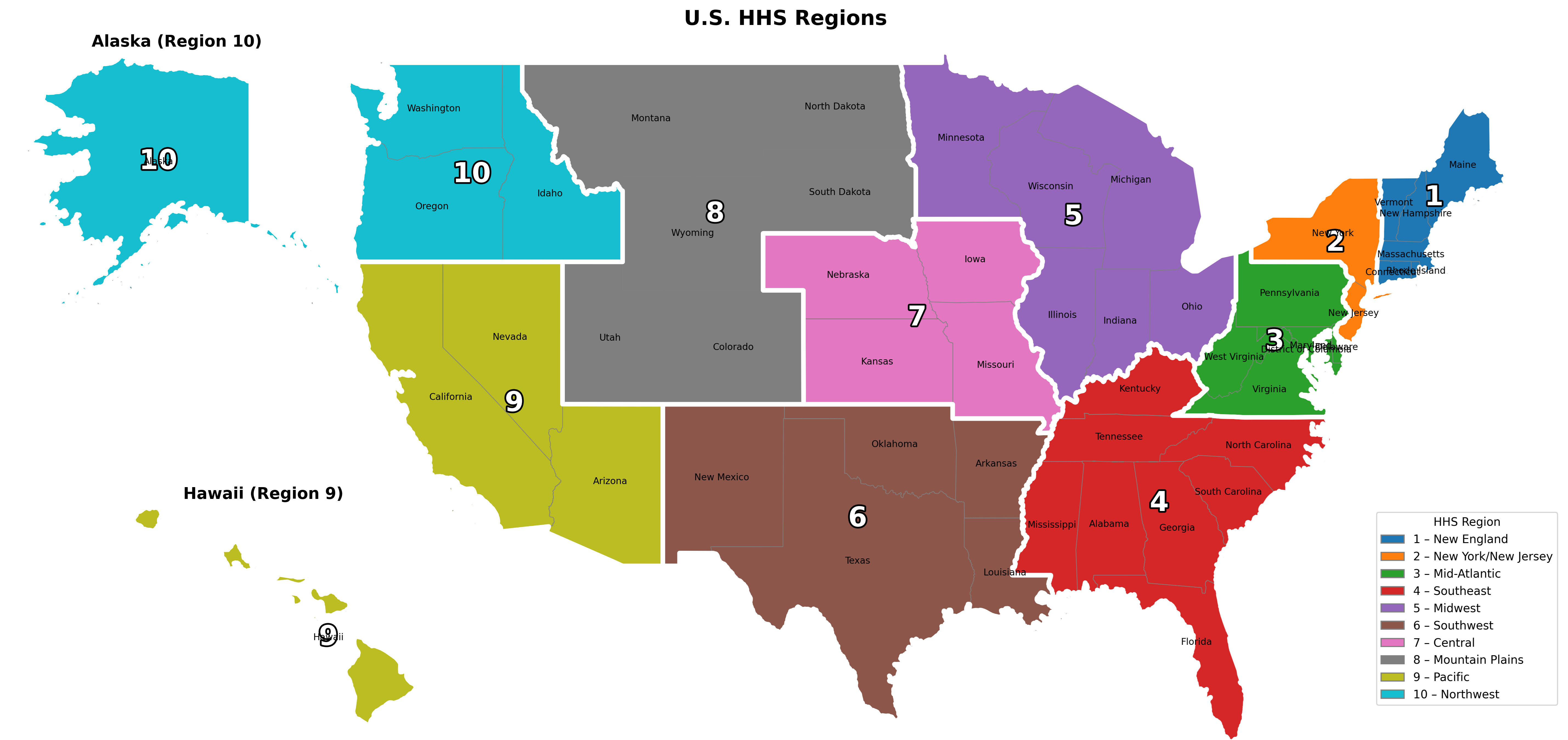}
    \caption{U.S. HHS regions used for spatial evaluation.}
    \label{fig:us_hhs_map}
  \end{subfigure}\hfill
  \begin{subfigure}[t]{0.45\textwidth}
    \centering
    \includegraphics[width=\linewidth]{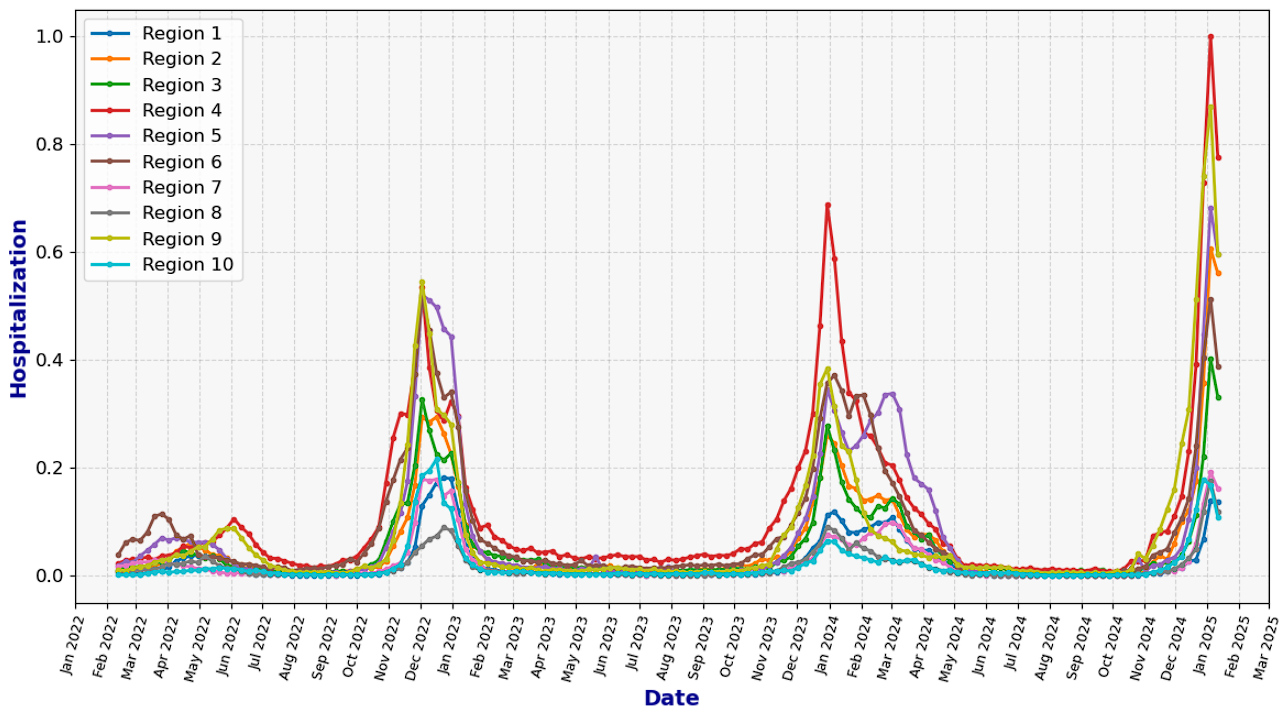}
    \caption{Weekly influenza hospitalization time series across regions.}
    \label{fig:hospital_data}
  \end{subfigure}
  \medskip
  % Bottom row: full-width
  \begin{subfigure}[b]{0.95\textwidth}
    \centering
    \includegraphics[width=\linewidth]{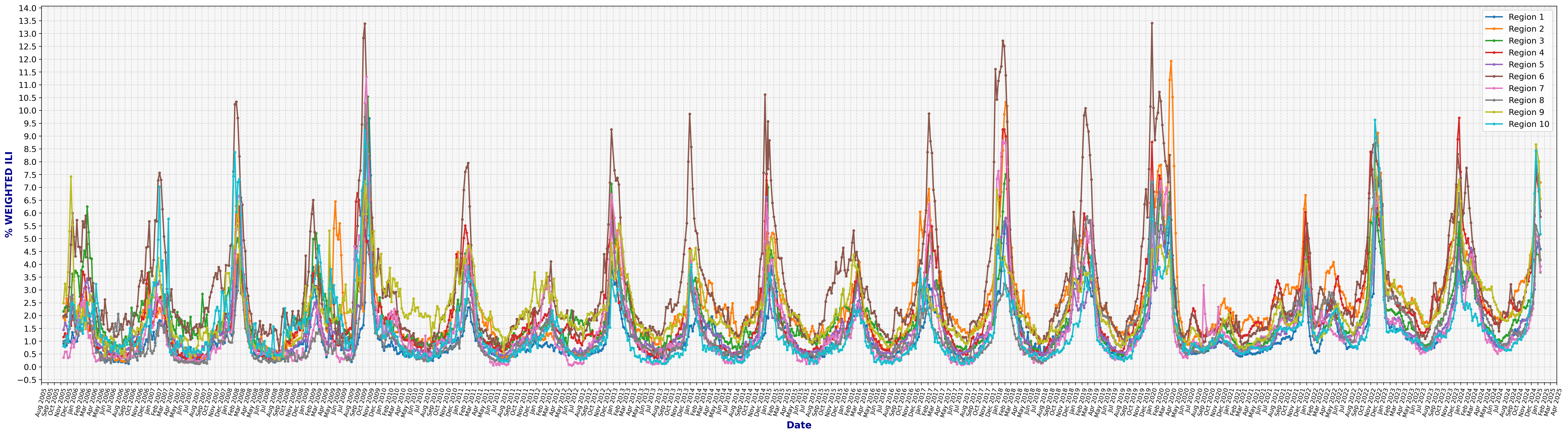} 
    \caption{Weekly influenza-like illness (ILI) time series across regions.}
    \label{fig:regions_series}
  \end{subfigure}

  \caption{Overview of the regional influenza forecasting data.}
  \label{fig:overview}
\end{figure*}

Reliable evaluation in epidemic forecasting hinges on transparent, leakage-free data handling across targets and pretraining corpora. We therefore standardize sources, calendar alignment, and revision policies, and we explicitly separate \emph{evaluation targets} (ILI and influenza hospitalizations) from \emph{auxiliary pretraining} datasets (TrafficL~\cite{godahewa2021monash}, M4~\cite{makridakis2020m4}, and other epidemic datasets). This section details each dataset, our two generalization regimes (temporal and spatial), and the preprocessing steps applied uniformly across models.

\paragraph{Influenza-Like Illness (ILI)}
Our primary evaluation uses weekly Influenza-Like Illness (ILI) time series from the CDC’s ILINet surveillance system. ILINet reports the percentage of outpatient visits attributable to fever plus cough or sore throat without another known cause. We analyze roughly twenty years of weekly observations per U.S. Health and Human Services (HHS) region (10 regions), extending into early 2025. Each regional series (on the order of $1{,}000$ points) exhibits strong seasonality, interannual variability, and pandemic-era deviations, making it a robust benchmark for real-time public-health forecasting. The processed regional ILI dataset and related repository materials are publicly available online.\footnote{\url{https://github.com/alireza-jafari/ILI-Influenza-Dataset}}

\paragraph{Influenza-Associated Hospitalizations}
We additionally evaluate weekly region-level counts of laboratory-confirmed influenza-associated hospitalizations (per $100{,}000$ population). In our experiments, this series spans approximately the past three influenza seasons and serves two roles: (i) a second target for out-of-sample validation, and (ii) an auxiliary signal to test whether models trained on ILI can also transfer to related but more severe influenza outcomes, such as hospitalizations.

While the Influenza-related datasets form the basis of all evaluation results, several additional datasets are incorporated at earlier stages of model development as auxiliary sources for pre-training or feature augmentation. To initialize our time series foundation models and assess cross-domain transferability, we pretrain on the following datasets: (i) \textit{Epidemic Weekly Data} (COVID-19), comprising weekly region-level death counts by HHS region from March 2020 through early 2025, whose intervention-driven waves and reporting idiosyncrasies provide epidemic dynamics distinct from influenza; (ii) \textit{TrafficL} (Monash Repository), the California Department of Transportation lane-occupancy series at hourly cadence, offering high-frequency periodic structure and markedly different distributional properties from epidemiology—useful for testing whether time series foundation models (TSFMs) learn transferable temporal inductive biases; and (iii) \textit{M4}, a large, heterogeneous benchmark of $\sim$100{,}000 univariate series spanning multiple domains and sampling rates, employed as a generic pretraining source to capture broad time series regularities.

\subsection{Forecasting Regimes \& Preprocessing}

We consider two complementary generalization settings. In the \emph{temporal (within-region)} regime, each region’s time series is split chronologically into training, validation, and test segments; models are trained and tuned on the earlier segments and evaluated on disjoint, later time ranges from the \emph{same} region, thereby assessing forward-in-time performance without geographic shift. In the \emph{spatial (across-region)} regime, models are trained on a subset of regions and evaluated on entirely \emph{unseen} regions, isolating the ability to transfer across geographies with potentially different seasonal amplitudes, reporting practices, and demographic structure.

To ensure comparability and prevent information leakage, all targets are aligned to CDC report weeks, and any exogenous covariates (when used) are lagged to reflect their operational availability at the forecast origin. We adopt a strict revision policy in which each series is frozen to the version available as of the forecast date, disallowing retrospective access to backfilled values. For scale harmonization, we apply per-region robust normalization (median/IQR) fit exclusively on the training split and subsequently reused for validation and test. Finally, we construct sliding windows with look-back length $L$ and horizons $H \in \{1,2,3,4\}$ weeks in a manner consistent across all targets and models.

ILI and influenza hospitalizations serve as the \emph{evaluation} targets throughout the study. The Epidemic, TrafficL, and M4 corpora are employed for \emph{pretraining} time series foundation models and for probing cross-domain transferability. In selected experiments, we also examine cross-target transfer (e.g., pretraining on ILI and evaluating on hospitalizations, and conversely) to quantify the extent to which learned temporal structure generalizes across related epidemiological endpoints.

\section{Models Evaluation and Comparison}
\label{experiment}

Our study spans two public-health datasets; here, we open with the influenza-like illness (ILI) benchmark: four-week-ahead forecasting under a temporal split that withholds future weeks of the same regions. This setting mirrors real deployment—models must read seasonal cues from the past and stay calibrated as the lead time stretches from one to four weeks.

We evaluate every method with two complementary lenses. First is mean squared error (MSE), which penalizes large mistakes quadratically. Second is the normalized Nash–Sutcliffe efficiency (NNSE) \cite{nossent2012application}, a unit-scaled variant of Nash–Sutcliffe efficiency (NSE) with values in the range $[0,1]$. An NNSE of $1$ denotes a perfect match to the observed series, $0.5$ corresponds to the performance of predicting the historical mean. The formula for NNSE is:

\begin{align}
NSE = 1 - \frac{\sum_{i=1}^{n} (O_i - P_i)^2}{\sum_{i=1}^{n} (O_i - \bar{O})^2} \\
NNSE = 1 / (2-NSE)
\end{align}

where \( O_i \) is the observed value at time \( i \), \( P_i \) is the predicted value at time \( i \), \( \bar{O} \) is the mean of the observed values, \( n \) is the number of observations. 

\subsection{Influenza-like illness (ILI) Prediction}

%------------------------------------------------------------------

\subsubsection{\textbf{Temporal evaluation: within-region forecasting}} 

With this frame in place, we now present the 1$-$4 week horizon ILI forecast evaluation results in Table \ref{tab:temporal_model_vs_horizon_mse_nnse}. We designed this experiment for several reasons. First, we wanted a clean, apples-to-apples comparison between foundation models and pattern models for influenza prediction. Foundation models are pretrained on large, related time series corpora (e.g., hospitalizations, M4, TrafficL) and then lightly fine-tuned to ILI, so they test whether broad temporal priors transfer to public-health forecasting. Pattern models are trained/fine-tuned directly on ILI without cross-domain pretraining, so they reveal how far domain-specific architectures can go on task data alone. This contrast matters because real-time deployment often balances two challenges: leverage whatever prior signal is available vs. keep pipelines simple and task-focused. Second, we explicitly compare LLM-based time series models (e.g., Chronos variants, TimeLLM), which tokenize or embed series through language-style pipelines, with numerical time series models (e.g., PatchTST, iTransformer, etc.) that operate directly on numeric sequences. This separation is important because LLM-style pretraining promises broad generalization, but numerical architectures encode inductive biases, such as, seasonality, multi-scale patterns, known to matter for epidemics. Understanding when language-style pretraining helps (and when it doesn’t) is essential before relying on it for operational forecasts. Third, we contrast iterative vs. direct multi-output forecasting. Iterative forecast models predict one week ahead and roll forward, exposing how errors compound under realistic use; direct forecast models predict the entire 1–4-week vector at once, testing whether joint-horizon modeling stabilizes long-range accuracy.

\begin{table*}[!t]
    {\centering
    \small
    \caption{Temporal evaluation on ILI over 1--4-week-ahead forecasting horizons. Model performance is measured using MSE and NNSE, and “Type” denotes foundation (F), pattern (P), and statistical (S) architectures. The performance values are averaged over multiple independent runs. Superior model performance is characterized by a low MSE and an NNSE value close to 1. Detailed model configurations, training hyperparameters, and retained architecture defaults are reported in Appendix Tables~\ref{tab:appendix_table_i_model_configs} and~\ref{tab:appendix_architecture_defaults_all_models}.}
    \label{tab:temporal_model_vs_horizon_mse_nnse}

    \vspace{-0.4em}

    \begin{adjustbox}{max width=\textwidth}
    \begin{tabular}{lll>{\bfseries}c>{\bfseries}ccccccccc}
        \toprule
        & & & \multicolumn{2}{c}{\textbf{Avg.\ (All 1--4)}} &
          \multicolumn{2}{c}{\textbf{1-week}} & \multicolumn{2}{c}{\textbf{2-week}} &
          \multicolumn{2}{c}{\textbf{3-week}} & \multicolumn{2}{c}{\textbf{4-week}} \\
        \cmidrule(lr){4-5}\cmidrule(lr){6-7}\cmidrule(lr){8-9}\cmidrule(lr){10-11}\cmidrule(lr){12-13}
        \textbf{Model} & \hspace{-3mm}\textbf{Type} & \textbf{Strategy} & MSE & NNSE & MSE & NNSE & MSE & NNSE & MSE & NNSE & MSE & NNSE \\
        \midrule
        ARIMA                           & S & Iterative & 0.962 & 0.739 & 0.145 & 0.938 & 0.558 & 0.803 & 1.163 & 0.669 & 1.981 & 0.546 \\
        Chronos-Bolt-mini               & F & Iterative & 0.764 & 0.772 & 0.201 & 0.922 & 0.561 & 0.810 & 0.939 & 0.718 & 1.355 & 0.638 \\
        Chronos-Bolt-small              & F & Iterative & 0.747 & 0.775 & 0.193 & 0.925 & 0.560 & 0.810 & 0.917 & 0.722 & 1.317 & 0.644 \\
        TimesNet                        & P & Direct    & 0.737 & 0.775 & 0.278 & 0.896 & 0.579 & 0.805 & 0.934 & 0.721 & 1.155 & 0.676 \\
        Chronos-Bolt-base               & F & Iterative & 0.730 & 0.780 & 0.183 & 0.929 & 0.534 & 0.817 & 0.898 & 0.726 & 1.305 & 0.646 \\
        TimeLLM-GPT2                    & F & Direct    & 0.724 & 0.772 & 0.394 & 0.858 & 0.617 & 0.795 & 0.835 & 0.741 & 1.050 & 0.695 \\
        Chronos-T5-large                & F & Iterative & 0.719 & 0.784 & 0.153 & 0.940 & 0.505 & 0.825 & 0.901 & 0.726 & 1.315 & 0.645 \\
        Chronos-T5-mini                 & F & Iterative & 0.716 & 0.784 & 0.166 & 0.935 & 0.507 & 0.825 & 0.902 & 0.726 & 1.287 & 0.650 \\
        Chronos-T5-base                 & F & Iterative & 0.683 & 0.791 & 0.151 & 0.940 & 0.483 & 0.831 & 0.850 & 0.737 & 1.247 & 0.657 \\
        TiDE                            & P & Direct    & 0.591 & 0.810 & 0.183 & 0.929 & 0.457 & 0.839 & 0.742 & 0.763 & 0.982 & 0.708 \\
        TCN                             & P & Direct    & 0.587 & 0.807 & 0.322 & 0.883 & 0.494 & 0.829 & 0.661 & 0.783 & 0.869 & 0.733 \\
        VanillaTransformer              & P & Direct    & 0.548 & 0.820 & 0.202 & 0.922 & 0.448 & 0.842 & 0.644 & 0.788 & 0.898 & 0.727 \\
        iTransformer-Epidemic           & F & Direct    & 0.490 & 0.836 & 0.153 & 0.940 & 0.379 & 0.863 & 0.601 & 0.799 & 0.827 & 0.743 \\
        LSTM-iterative                  & P & Iterative & 0.478 & 0.837 & 0.121 & 0.947 & 0.358 & 0.864 & 0.605 & 0.796 & 0.829 & 0.740 \\
        iTransformer                    & P & Direct    & 0.471 & 0.841 & 0.162 & 0.936 & 0.370 & 0.866 & 0.569 & 0.807 & 0.784 & 0.753 \\
        iTransformer-TrafficL           & F & Direct    & 0.465 & 0.843 & 0.147 & 0.942 & 0.362 & 0.868 & 0.568 & 0.808 & 0.784 & 0.753 \\
        TFT                             & P & Direct    & 0.454 & 0.845 & 0.163 & 0.936 & 0.366 & 0.867 & 0.545 & 0.811 & 0.734 & 0.764 \\
        LSTM-direct                     & P & Direct    & 0.449 & 0.844 & 0.145 & 0.938 & 0.350 & 0.867 & 0.564 & 0.807 & 0.735 & 0.765 \\
        iTransformer-M4                 & F & Direct    & 0.443 & 0.849 & 0.143 & 0.944 & 0.349 & 0.872 & 0.543 & 0.815 & 0.736 & 0.764 \\
        PatchTST                        & P & Direct    & 0.439 & 0.850 & 0.129 & 0.949 & 0.355 & 0.870 & 0.548 & 0.813 & 0.722 & 0.767 \\
        iTransformer-Hospitalization    & F & Direct    & 0.422 & 0.855 & 0.138 & 0.945 & 0.330 & 0.878 & 0.518 & 0.822 & 0.702 & 0.773 \\
        PatchTST-TrafficL               & F & Direct    & 0.422 & 0.855 & 0.131 & 0.948 & 0.345 & 0.874 & 0.527 & 0.819 & 0.684 & 0.777 \\
        PatchTST-Epidemic               & F & Direct    & 0.419 & 0.855 & 0.133 & 0.947 & 0.335 & 0.877 & 0.523 & 0.820 & 0.686 & 0.777 \\
        PatchTST-M4                     & F & Direct    & 0.415 & 0.856 & 0.130 & 0.948 & 0.337 & 0.876 & 0.517 & 0.822 & 0.677 & 0.779 \\
        PatchTST-Hospitalization        & F & Direct    & 0.412 & 0.857 & 0.124 & 0.950 & 0.331 & 0.878 & 0.515 & 0.822 & 0.678 & 0.779 \\
        \textbf{MultiFoundationCore}    & F & Direct    & 0.382 & 0.864 & 0.118 & 0.949 & 0.306 & 0.881 & 0.475 & 0.832 & 0.627 & 0.792 \\
        \bottomrule
    \end{tabular}
    \end{adjustbox}
    }

    {\normalsize
    \vspace{0.75em}
    \refstepcounter{figure}
    \label{fig:temporal_ili_h1_h4}
    \includegraphics[width=1.0\textwidth]{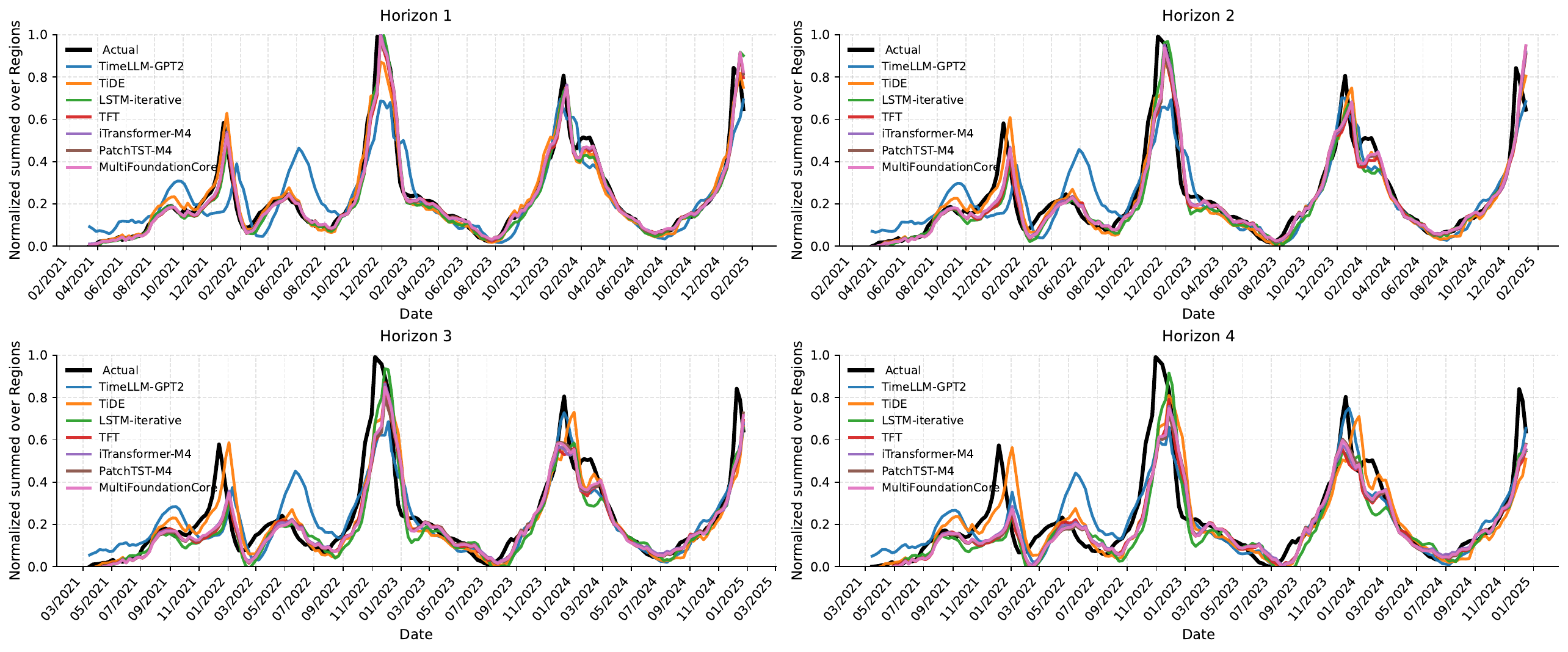}
    \vspace{-0.05em}
    \noindent
    \textbf{Fig.~\thefigure.}
    Temporal evaluation on ILI over 1--4-week-ahead forecasting horizons. Normalized sum over regions, comparing model predictions with the observed values over the test data.
}
\end{table*}

Several clear patterns emerge from Table~\ref{tab:temporal_model_vs_horizon_mse_nnse}. We start with MultiFoundationCore, which is a mixture-of-experts model. It achieves the best overall average on the ILI temporal split (Avg. MSE = 0.382; NNSE = 0.864). In short, ensembling pretrained experts with task-aware fusion produces forecasts that stay calibrated as the horizon stretches.

The PatchTST variants provide a clear example of the value of pretraining for epidemic forecasting. PatchTST without pretraining achieves Avg. 0.439 MSE / 0.850 NNSE. Pretraining on Hospitalization (closely related epidemiological signal) improves the average to 0.412 / 0.857; M4 (very large, diverse time series corpus) yields 0.415 / 0.856; Epidemic dataset gives 0.419 / 0.855; and TrafficL (non-epidemiological but structured) gives 0.422 / 0.855. The pattern is consistent and instructive: (i) Hospitalization transfers best on average—reasonable given its mechanistic link to ILI; (ii) M4 helps especially at the longest horizons (e.g., week-4 MSE = 0.677, the best 4-week single-model score), likely because its scale and variety teach robust seasonal and trend templates; and (iii) Epidemic and TrafficL also help, confirming that large-scale temporal priors—even from adjacent domains—stabilize multi-week ILI forecasts. Net: PatchTST is a perfect fit for this problem, and pretraining makes it better, then better again.

Transformer exhibits a similar transfer-learning trend. As a pattern model, iTransformer averages 0.471 / 0.841. Pretraining on Hospitalization and M4 cuts that to 0.422 / 0.855 and 0.443 / 0.849, respectively; TrafficL also helps (0.465 / 0.843). Taken together, iTransformer mirrors PatchTST’s behavior: pretraining almost always helps, with the strongest gains coming from related (Hospitalization) or very large (M4) corpora.

Among pattern models, TFT is a strong, stable baseline (0.454 / 0.845) and degrades gently through week 4, reflecting its seasonality/temporal-fusion inductive biases. The LSTM comparison illustrates horizon strategy: the iterative LSTM is highly competitive at week 1 (MSE = 0.121) but degrades by week 4 (MSE = 0.829), whereas the direct multi-output LSTM curbs compounding and delivers a better average (0.449 / 0.844). 

LLM-style approaches lag here. The Chronos family and TimeLLM—which bring language-style pretraining and tokenization to time series—do not match the best numerical architectures. The strongest Chronos configuration (T5-base) averages 0.683 / 0.791. TimeLLM averages 0.724 / 0.772, despite claims of strong ILI performance in its paper \cite{jin2024timellmtimeseriesforecasting}. In part, this seems tied to mismatch in horizon design: TimeLLM emphasizes very long or non-standard horizons that are less aligned with CDC’s 1–4-week operational focus; when evaluated under our operationally relevant 1–4-week regime, its advantage disappears.

Table~\ref{tab:temporal_model_vs_horizon_mse_nnse} also clarifies how forecasting strategy interacts with the multi-horizon setting. The top-performing rows in Table~\ref{tab:temporal_model_vs_horizon_mse_nnse} are almost exclusively direct multi-output forecasters: they estimate the full 1–4 week trajectory in a single pass and therefore avoid reusing their own predictions as inputs. In contrast, iterative models—especially the LSTM variants—are competitive or even excellent at the first week but lose ground as the horizon increases. For example, LSTM-iterative is ranked second-best at 1-week, yet its overall rank drops to 13 once weeks 2–4 are included, while the direct LSTM version stays in the top ten across all horizons. This pattern is consistent with the expected accumulation of error: iterative predictors repeatedly feed their own noisy outputs back into the model, so small short-term mistakes can grow into substantial long-term deviations, which is precisely what the rank degradation at weeks 3–4 reflects.

The PatchTST variants provide a more fine-grained view of horizon-dependent behavior under different pretraining sources. PatchTST-Hospitalization is very strong and stable across all horizons (ranks 3, 3, 2, and 3), while PatchTST-M4 starts slightly lower at week 1 (rank 5) but rises to rank 2 by week 4. This suggests that M4 pretraining is particularly helpful for the longest lead time, where broad seasonal and trend structure learned from a large, heterogeneous corpus becomes most valuable. PatchTST-Epidemic shows a different profile: it is weaker at week 1 (rank 7) but improves to rank 5 by week 4, indicating moderate gains that are more evenly spread across horizons. In contrast, the ILI-only PatchTST baseline begins very strong at week 1 (rank 4) yet falls to rank 7 at week 4, highlighting that models trained solely on task data can excel in the short term but lack the long-range robustness conferred by cross-domain pretraining.

%------------------------------------------------------------------

Table~\ref{tab:model_results_one_standalone} zooms in on the 1-week horizon as a standalone forecasting task. This isolates pure short-horizon skill without any trade-offs from longer leads or error propagation. As expected, almost all models improve their 1-week MSE and NNSE compared to their week-1 performance in the multi-week setting, but the magnitude of these gains—and how they differ across architectures—is informative.

\begin{table}[ht!]
    \label{table1}
    \centering
    \small
    \caption{Temporal evaluation on ILI over a standalone 1-week-ahead forecasting horizon. “Pre-trained/Fine-tuned” column indicates whether each model is initialized from a pretrained checkpoint and/or fine-tuned on ILI.}
    \label{tab:model_results_one_standalone}
    \begin{adjustbox}{max width=\textwidth}
    \begin{tabular}{lccccc}
        \toprule
        \textbf{Model} & \textbf{Type} & \textbf{PT/FT} & \textbf{MSE} & \textbf{NNSE} \\
        \midrule
        TimeLLM-GPT2          & F & Yes/No & 10.429 & 0.228 \\
        TimeLLM-GPT2          & F & Yes/Yes & 0.337  & 0.876 \\
        TimesNet              & P & No/Yes & 0.215  & 0.917 \\
        Chronos-Bolt-mini     & F & Yes/No & 0.200  & 0.922 \\
        Chronos-Bolt-base     & F & Yes/No & 0.193  & 0.925 \\
        Chronos-Bolt-small    & F & Yes/No & 0.182  & 0.928 \\
        VanillaTransformer    & P & No/Yes & 0.163  & 0.935 \\
        TCN                   & P & No/Yes & 0.158  & 0.937 \\
        Chronos-T5-mini       & F & Yes/No & 0.158  & 0.937 \\
        TiDE                  & P & No/Yes & 0.154  & 0.939 \\
        Chronos-T5-large      & F & Yes/No & 0.153  & 0.939 \\
        Chronos-T5-base       & F & Yes/No & 0.151  & 0.940 \\
        TFT                   & P & No/Yes & 0.143  & 0.943 \\
        iTransformer          & P & No/Yes & 0.132  & 0.947 \\
        PatchTST              & P & No/Yes & 0.127  & 0.949 \\
        LSTM-iterative        & P & No/Yes & 0.122  & 0.950 \\
        \textbf{MultiFoundationCore}   & F & Yes/Yes & 0.112 & 0.954 \\
        \bottomrule
    \end{tabular}
    \end{adjustbox}
\end{table}

MultiFoundationCore remains the strongest model under the single-step regime. Notably, LSTM emerges as the second-best model, which further highlights how strong this prediction setting is even for comparatively simpler architectures. PatchTST improves from 0.129 to 0.127 MSE, and iTransformer from 0.162 down to 0.132. TFT, TiDE, TCN, and VanillaTransformer show similar trends, with TCN in particular nearly halving its 1-week error relative to the multi-horizon setup. These shifts indicate that some architectures—especially convolutional and transformer baselines—sacrifice near-term accuracy when they must cover weeks 2–4 simultaneously, but recover strong short-range performance when trained solely for week 1. 

At the same time, the relative ordering of the best numeric models largely persists: PatchTST, LSTM, and iTransformer still form a tight top cluster behind MultiFoundationCore, suggesting that the inductive biases that mattered in the CDC-aligned 1–4-week regime continue to matter when we focus exclusively on the first week. Importantly, iterative models are strong in 1-week forecasting: LSTM-iterative is the second-best model, and Chronos models are also competitive here. However, in the multi-week evaluation, Chronos falls into the poorer group on average, while MultiFoundationCore stays best because it explicitly uses these forecasters’ outputs as input streams/signals and learns to aggregate them into a more robust prediction.

%------------------------------------------------------------------

\subsubsection{\textbf{Spatial evaluation: across-region generalization}} 

To complete our study of the influenza forecasting problem, we also evaluate models under a \emph{spatial} split, where training and test sets consist of disjoint regions. This setting probes across-region generalization rather than purely temporal extrapolation within a region. We consider two configurations: (i) a standalone 1-week-ahead task (Table~\ref{tab:spatial_model_results}), and (ii) a 4-week multi-output setting (Table~\ref{tab:spatial_model_vs_horizon_mse_nnse}) analogous to the temporal experiment in Table~\ref{tab:temporal_model_vs_horizon_mse_nnse}. In this section we restrict attention to the strongest architectures from the temporal results (PatchTST, LSTM, TFT, and representative foundation baselines), and we do not include \emph{MultiFoundationCore} because its design is inherently temporal and is not compatible with a clean across-region test.

\begin{table}[ht!]
    \label{table2}
    \centering
    \small
    \caption{Spatial evaluation on ILI for a standalone 1-week-ahead forecasting horizon. “Pre-train/Fine-tune” indicates whether models are initialized from pretrained checkpoints and/or fine-tuned on ILI.}
    \label{tab:spatial_model_results}
    \begin{adjustbox}{max width=\textwidth}
    \begin{tabular}{lccccc}
        \toprule
        \textbf{Model} & \textbf{Type} & \textbf{PT/FT} & \textbf{MSE} & \textbf{NNSE} \\
        \midrule
        TimeLLM-GPT2          & F & Yes/No & 10.839 & 0.239 \\
        TiDE                  & P & No/Yes & 0.392  & 0.861 \\
        TimeLLM-GPT2          & F & Yes/Yes & 0.357  & 0.872 \\
        TimesNet              & P & No/Yes & 0.336  & 0.885 \\
        iTransformer          & P & No/Yes & 0.256  & 0.904 \\
        Chronos-Bolt-small    & F & Yes/No & 0.254  & 0.910 \\
        Chronos-Bolt-mini     & F & Yes/No & 0.253  & 0.910 \\
        Chronos-Bolt-base     & F & Yes/No & 0.247  & 0.912 \\
        TCN                   & P & No/Yes & 0.243  & 0.912 \\
        VanillaTransformer    & P & No/Yes & 0.214  & 0.920 \\
        Chronos-T5-large      & F & Yes/No & 0.214  & 0.923 \\
        Chronos-T5-mini       & F & Yes/No & 0.213  & 0.923 \\
        Chronos-T5-base       & F & Yes/No & 0.218  & 0.922 \\
        LSTM                  & P & No/Yes & 0.197  & 0.926 \\
        TFT                   & P & No/Yes & 0.191  & 0.927 \\
        \textbf{PatchTST}     & P & No/Yes & 0.166  & 0.935 \\
        \bottomrule
    \end{tabular}
    \end{adjustbox}
\end{table}

Table~\ref{tab:spatial_model_results} reports 1-week-ahead spatial performance. Pattern models remain dominant: PatchTST is the best model (MSE $=0.166$, NNSE $=0.935$), followed by TFT (MSE $=0.191$, NNSE $=0.927$) and LSTM (MSE $=0.197$, NNSE $=0.926$). Architectures such as TCN, TiDE, TimesNet, and iTransformer form a strong middle tier. Compared to the temporal 1-week table, all models incur modestly higher error under spatial splitting (e.g., PatchTST: MSE $0.127 \rightarrow 0.166$; LSTM: $0.122 \rightarrow 0.197$), reflecting the added difficulty of generalizing to unseen regions.

Chronos remains competitive in the 1-week spatial setting, outperforming TimeLLM-GPT2. Chronos is a language-style pretraining foundation model that treats numerical time series values as discrete tokens through quantization. By contrast, TimeLLM-GPT2 showed weaker performance in our experiments. This distinction suggests that language-style time-series models do not fail uniformly: Chronos remains competitive in short-horizon settings, while TimeLLM’s weaker results support the concern that language backbones may not reliably preserve the fine-grained numerical and temporal structure needed for epidemic forecasting.

Table~\ref{tab:spatial_model_vs_horizon_mse_nnse} extends this analysis to 4-week-ahead spatial forecasting for the three strongest pattern models identified above. PatchTST again achieves the best average performance (Avg.\ MSE $=0.449$, NNSE $=0.848$), followed by TFT (Avg.\ MSE $=0.502$, NNSE $=0.836$) and LSTM (Avg.\ MSE $=0.609$, NNSE $=0.812$). As the horizon increases from 1 to 4 weeks, all three models show rising MSE and decreasing NNSE, with the sharpest degradation for LSTM. This pattern is consistent with the temporal 4-week experiment: architectures with strong inductive biases for multi-scale temporal structure (PatchTST, TFT) remain more stable at longer horizons than a plain recurrent baseline.

Using the scale-normalized NNSE metric makes the trade-off between temporal and spatial evaluation particularly clear. Under the temporal split, the best model, MultiFoundationCore, achieves an average NNSE of $0.864$ (Table~\ref{tab:temporal_model_vs_horizon_mse_nnse}), whereas under the spatial split, the best model, PatchTST, reaches a slightly lower NNSE of $0.848$ (Table~\ref{tab:spatial_model_vs_horizon_mse_nnse}). This comparison indicates that, for this dataset, models generally perform better under temporal splitting than under spatial splitting. A likely reason is that the dataset contains only 10 regions, which limits the diversity available for cross-region generalization. With a larger number of time series drawn from smaller and more diverse regions, spatial forecasting performance might improve.

\begin{table*}[ht!]
    \centering
    \small
    \caption{Spatial evaluation on ILI over 1–4-week-ahead forecasting horizons. Model performance is measured using MSE and NNSE, and “Type” denotes foundation (F) and pattern (P) architectures.}
    \label{tab:spatial_model_vs_horizon_mse_nnse}
    \begin{adjustbox}{max width=\textwidth}
    \begin{tabular}{llcccccccccc}
        \toprule
        & & \multicolumn{2}{c}{\textbf{Avg.\ (All 1--4)}} & \multicolumn{2}{c}{\textbf{1-week}} & \multicolumn{2}{c}{\textbf{2-week}} &
        \multicolumn{2}{c}{\textbf{3-week}} & \multicolumn{2}{c}{\textbf{4-week}} \\
        \cmidrule(lr){3-4}\cmidrule(lr){5-6}\cmidrule(lr){7-8}\cmidrule(lr){9-10}\cmidrule(lr){11-12}
        \textbf{Model} & \hspace{-3mm}\textbf{Type} & MSE & NNSE & MSE & NNSE & MSE & NNSE & MSE & NNSE & MSE & NNSE \\
        \midrule
        LSTM     & P & \textbf{0.609} & \textbf{0.812} & 0.205 & 0.925 & 0.473 & 0.841 & 0.754 & 0.768 & 1.003 & 0.715 \\
        TFT      & P & \textbf{0.502} & \textbf{0.836} & 0.217 & 0.920 & 0.426 & 0.853 & 0.613 & 0.802 & 0.751 & 0.768 \\
        PatchTST & P & \textbf{0.449} & \textbf{0.848} & 0.165 & 0.936 & 0.366 & 0.868 & 0.555 & 0.814 & 0.709 & 0.775 \\
        \bottomrule
    \end{tabular}
    \end{adjustbox}
\end{table*}

%------------------------------------------------------------------

\subsubsection{\textbf{Retraining frequency and temporal adaptation}}

In all previous temporal experiments, each model was trained once on a fixed history and then evaluated on a held-out test window. In practice, however, public-health forecasters routinely retrain or update their models as new data arrive. To study how such temporal adaptation affects performance on the ILI task, we conduct a targeted experiment with PatchTST where the model is periodically retrained on all data available up to the current time. We control the retraining frequency with a single parameter, \texttt{retrain\_window (rw)}: after every \texttt{rw} prediction steps, PatchTST is refit from scratch on the full ILI history observed so far, and then used to generate the next block of multi-horizon forecasts. Table~\ref{tab:patchtst_wr_vs_horizon_mse_nnse} reports the resulting 1–4-week errors for \texttt{rw} $\in \{200, 100, 50, 10\}$ under the same temporal split as in our main experiments.

The results show that more frequent retraining yields consistent, though gradually diminishing, gains in multi-week accuracy. With the least adaptive setting (\texttt{rw} = 200), PatchTST achieves an average MSE of 0.439 and NNSE of 0.850 across the 1–4-week horizons. Reducing the retraining window to 100 and 50 samples progressively improves the average MSE and NNSE. The most adaptive configuration (\texttt{rw} = 10) attains the best overall performance, with average MSE 0.422 and NNSE 0.854. 

From a computational standpoint, retraining at every single time step would be prohibitively expensive. Retraining every 200 steps leaves some performance on the table. Moving to \texttt{rw} = 10 increases computational cost, and the table shows that \texttt{rw} = 10 almost saturates the accuracy curve: the improvement from \texttt{rw} = 50 to \texttt{rw} = 10 is modest compared to the jump from \texttt{rw} = 200 to 100. This suggests that for ILI forecasting, a moderate retraining frequency offers a good balance between computational burden and predictive performance. This observation is consistent with a broader view that temporal choices are not merely experimental details, but can significantly affect the performance of models in biomedical prediction problems~\cite{jafari2026temporal}.

\begin{table*}[ht!]
    \centering
    \small
    \caption{Temporal ILI evaluation of PatchTST under different retraining frequencies, controlled by \texttt{retrain\_window} (rw).}
    \label{tab:patchtst_wr_vs_horizon_mse_nnse}
    \begin{adjustbox}{max width=\textwidth}
    \begin{tabular}{lcccccccccc}
        \toprule
        & \multicolumn{2}{c}{\textbf{Avg.\ (All 1--4)}} & \multicolumn{2}{c}{\textbf{1-week}} & \multicolumn{2}{c}{\textbf{2-week}} &
          \multicolumn{2}{c}{\textbf{3-week}} & \multicolumn{2}{c}{\textbf{4-week}} \\
        \cmidrule(lr){2-3}\cmidrule(lr){4-5}\cmidrule(lr){6-7}\cmidrule(lr){8-9}\cmidrule(lr){10-11}
        \textbf{Configuration} & MSE & NNSE & MSE & NNSE & MSE & NNSE & MSE & NNSE & MSE & NNSE \\
        \midrule
        $\texttt{rw}=200$ & \textbf{0.439} & \textbf{0.850} & 0.129 & 0.949 & 0.355 & 0.870 & 0.548 & 0.813 & 0.722 & 0.767 \\
        $\texttt{rw}=100$ & \textbf{0.427} & \textbf{0.853} & 0.131 & 0.948 & 0.345 & 0.874 & 0.534 & 0.817 & 0.698 & 0.774 \\
        $\texttt{rw}=50$  & \textbf{0.424} & \textbf{0.854} & 0.130 & 0.948 & 0.343 & 0.874 & 0.529 & 0.818 & 0.695 & 0.774 \\
        $\texttt{rw}=10$  & \textbf{0.422} & \textbf{0.854} & 0.130 & 0.948 & 0.346 & 0.873 & 0.526 & 0.819 & 0.684 & 0.777 \\
        \bottomrule
    \end{tabular}
    \end{adjustbox}
\end{table*}

%------------------------------------------------------------------

\subsubsection{\textbf{Effect of influenza hospitalizations on ILI prediction}}

Table~\ref{tab:new_temporal_all_models} probes how much domain history and related-domain signal matter for multi-horizon ILI forecasting, and whether those signals are better used as pre-training or as additional inputs at inference time. All metrics are computed on ILI; hospitalization is never a target here—it is either a source for pre-training or an auxiliary covariate (“input stream”). We focus on a six-month temporal evaluation window at the end of the series and vary both (i) how much ILI history is available (3 vs.\ 20 years) and (ii) how hospitalization is used: not at all, as a pre-training domain, as an exogenous input stream, or both. This design isolates three questions: does a longer ILI archive help; does transfer from hospitalization help beyond that, and how far can a hybrid, foundation-model–aided approach push the error floor.

A key limitation of current time series foundation models, such as TimeLLM, in their standard form, is that they are effectively single-target forecasters: they cannot natively ingest rich epidemiological covariates as additional input channels. The only way for them to use hospitalization information is via pre-training on hospitalization series and then fine-tuning on ILI. By contrast, our MultiFoundationCore architecture is explicitly structured to handle both pre-trained components \emph{and} auxiliary input streams. This experiment therefore highlights not only the value of hospitalization data but also the importance of architectures that can exploit auxiliary covariates.

Hospitalization data contributes a complementary signal in two distinct ways. When used as an auxiliary input stream alongside short ILI histories, it substantially lifts accuracy and stabilizes NNSE at longer horizons. For example, with only three years of ILI, adding hospitalization as an input stream reduces TFT’s average MSE. This suggests that contemporaneous hospitalization covariates help disambiguate trend and seasonal phase. When hospitalization is used as a pre-training domain for the forecaster, the benefits are larger and more consistent across all horizons. For TFT, adding hospitalization pre-training on top of twenty years of ILI reduces the average MSE from 0.428 to 0.346 and raises NNSE from 0.849 to 0.873; for PatchTST, the average MSE drops from 0.363 to 0.315 with NNSE increasing from 0.870 to 0.883. Improvements are modest at one week but widen at weeks 3–4, indicating that cross-domain transfer helps the model internalize slower-moving, clinically mediated signals that drive medium-range dynamics.

TimeLLM under this pre-training setup underperforms pattern numeric architectures in this setting (average MSE 0.638, NNSE 0.783), reinforcing that token-centric LLMs require careful task-specific adaptation to approach competitive numerical forecasting, especially as the horizon lengthens. In contrast, the MoE-style MultiFoundationCore setup sets a new best across all horizons. With hospitalization used only as a pre-training source, MFC attains an average MSE of 0.225 and NNSE of 0.915, substantially better than the strongest single-model baselines. When hospitalization is used both for pre-training and as an input stream at inference, the average MSE edges down further to 0.214, and NNSE rises to 0.917. The incremental gain from adding the input stream on top of pre-training is small but consistent, implying that most of the benefit is captured during representation learning, with a residual payoff from real-time covariate context.

Taken together, these results argue for three practical guidelines in ILI forecasting. First, longer, multi-season ILI histories are disproportionately valuable for the multi-week horizons that matter for planning. Second, cross-domain transfer from hospitalization helps most when embedded as pre-training, and its impact grows with the forecast horizon. Third, fusing pretrained foundation models with pattern-specialist forecasters in an architecture that can also ingest hospitalization as an exogenous input yields the strongest and most stable performance, pushing errors down across one- to four-week look-aheads while maintaining high NNSE—precisely the regime where decision-makers need reliable lift.

\begin{table*}[ht!]
    \centering
    \small
    \caption{Temporal ILI evaluation with hospitalization as a related-domain signal: mean MSE and NNSE for 1–4-week-ahead forecasts under different uses of ILI history (3 vs.\ 20 years) and hospitalization data (none, pre-training corpus, input stream, or both). All metrics are computed on ILI.}
    \label{tab:new_temporal_all_models}
    \begin{adjustbox}{max width=\textwidth}
    \begin{tabular}{lcc  >{\bfseries}c>{\bfseries}c cc cc cc cc}
        \toprule
        & \multicolumn{1}{c}{} & \multicolumn{1}{c}{}
        & \multicolumn{2}{c}{\textbf{Avg.\ (All 1--4)}}
        & \multicolumn{2}{c}{\textbf{1-week}}
        & \multicolumn{2}{c}{\textbf{2-week}}
        & \multicolumn{2}{c}{\textbf{3-week}}
        & \multicolumn{2}{c}{\textbf{4-week}} \\
        \cmidrule(lr){4-5}
        \cmidrule(lr){6-7}
        \cmidrule(lr){8-9}
        \cmidrule(lr){10-11}
        \cmidrule(lr){12-13}
        \textbf{Model}
        & \textbf{ILI} & \textbf{Hospitalization as}
        & MSE & NNSE
        & MSE & NNSE
        & MSE & NNSE
        & MSE & NNSE
        & MSE & NNSE \\
        \midrule
        TimeLLM             & 20 years data   & pre-training              & 0.638 & 0.783 & 0.219 & 0.844 & 0.480 & 0.804 & 0.863 & 0.745 & 0.988 & 0.738 \\
        TFT                 & 3 years data    & --                        & 0.583 & 0.812 & 0.077 & 0.939 & 0.436 & 0.819 & 0.783 & 0.763 & 1.036 & 0.729 \\
        TFT                 & 3 years data    & input stream              & 0.486 & 0.833 & 0.085 & 0.933 & 0.396 & 0.832 & 0.653 & 0.794 & 0.810 & 0.774 \\
        TFT                 & 20 years data   & --                        & 0.428 & 0.849 & 0.099 & 0.923 & 0.312 & 0.863 & 0.554 & 0.820 & 0.746 & 0.789 \\
        PatchTST            & 20 years data   & --                        & 0.363 & 0.870 & 0.055 & 0.955 & 0.301 & 0.867 & 0.499 & 0.835 & 0.596 & 0.824 \\
        TFT                 & 20 years data   & pre-training              & 0.346 & 0.873 & 0.075 & 0.940 & 0.276 & 0.877 & 0.449 & 0.849 & 0.586 & 0.826 \\
        PatchTST            & 20 years data   & pre-training              & 0.315 & 0.883 & 0.056 & 0.955 & 0.279 & 0.876 & 0.423 & 0.856 & 0.504 & 0.847 \\
        MultiFoundationCore & 17/3 years data & pre-training              & 0.225 & 0.915 & 0.048 & 0.961 & 0.174 & 0.919 & 0.233 & 0.916 & 0.444 & 0.862 \\
        MultiFoundationCore & 17/3 years data & pre-training \& input     & 0.214 & 0.917 & 0.047 & 0.962 & 0.185 & 0.914 & 0.216 & 0.921 & 0.407 & 0.872 \\
        \bottomrule
    \end{tabular}
    \end{adjustbox}
\end{table*}

%------------------------------------------------------------------

\subsection{Influenza Hospitalization Prediction} 

In addition to forecasting ILI, we also evaluate models on predicting influenza-associated hospitalizations, the current gold-standard target in the FluSight influenza forecasting challenge~\cite{CDCFluSight2023}. This dataset differs from the ILI benchmark in two important ways. First, it covers a much shorter time span: we have only about three years of weekly hospitalization data, compared to up to twenty years of ILI history in our main experiments. Second, the raw hospitalization counts are much larger in magnitude. To avoid numerical issues and to make errors comparable across regions, we normalize the hospitalization series (per region) before training and evaluation. As in the ILI setting, we use a 4-week multi-horizon forecasting window aligned with CDC practice (1–4 weeks ahead). Given this limited data, we develop a compact \emph{TinyLSTM} baseline with only a few recurrent units, to test whether a small regularized model can remain competitive with higher-capacity architectures in short-data regimes.

\begin{table*}[ht!]
    \centering
    \small
    \caption{Temporal evaluation on normalized influenza-associated hospitalizations. The “Pre-trained dataset” column indicates whether each model is trained only on hospitalization histories (None) or additionally leverages ILI (as a pre-training domain and/or auxiliary signal).}
    \label{tab:temporal_model_vs_horizon_mse_ili}
    \begin{adjustbox}{max width=\textwidth}
    \begin{tabular}{lcccccc}
        \toprule
        \textbf{Model} & \textbf{Pre-trained dataset} & \textbf{Avg.\ (All 1--4)} & \textbf{1-week} & \textbf{2-week} & \textbf{3-week} & \textbf{4-week} \\
        \midrule
        TimeLLM-GPT2         & None & \textbf{0.00595} & 0.00144 & 0.00236 & 0.00819 & 0.01182 \\
        TimeLLM-GPT2         & ILI  & \textbf{0.00433} & 0.00033 & 0.00105 & 0.00621 & 0.00975 \\
        PatchTST             & None & \textbf{0.00361} & 0.00007 & 0.00057 & 0.00519 & 0.00859 \\
        iTransformer         & None & \textbf{0.00345} & 0.00007 & 0.00054 & 0.00500 & 0.00818 \\
        iTransformer         & ILI  & \textbf{0.00344} & 0.00006 & 0.00050 & 0.00500 & 0.00820 \\
        PatchTST             & ILI  & \textbf{0.00343} & 0.00006 & 0.00055 & 0.00498 & 0.00811 \\
        LSTM                 & None & \textbf{0.00329} & 0.00086 & 0.00115 & 0.00474 & 0.00641 \\
        TFT                  & None & \textbf{0.00320} & 0.00008 & 0.00048 & 0.00467 & 0.00758 \\
        TFT                  & ILI  & \textbf{0.00295} & 0.00007 & 0.00035 & 0.00429 & 0.00710 \\
        LSTM                 & ILI  & \textbf{0.00269} & 0.00019 & 0.00050 & 0.00417 & 0.00589 \\
        TinyLSTM             & None & \textbf{0.00262} & 0.00005 & 0.00033 & 0.00399 & 0.00611 \\
        TinyLSTM             & ILI  & \textbf{0.00237} & 0.00006 & 0.00029 & 0.00355 & 0.00560 \\
        \textbf{MultiFoundationCore}  & ILI  & \textbf{0.00224} & 0.00007 & 0.00033 & 0.00333 & 0.00523 \\
        \bottomrule
    \end{tabular}
    \end{adjustbox}
\end{table*}

Table~\ref{tab:temporal_model_vs_horizon_mse_ili} summarizes the mean MSE of several models under this four-horizon temporal setup, both \emph{with} and \emph{without} ILI as an auxiliary input. The “None” rows correspond to models that see only hospitalization histories; the “ILI” rows augment each model with the corresponding ILI series as an additional covariate. Even though the absolute MSE values are small due to normalization, the relative differences across models and configurations are informative. Among models that use only hospitalization, TinyLSTM achieves the lowest average error (0.00262). iTransformer and PatchTST remain competitive (0.00345 and 0.00361), while TimeLLM is clearly weaker in this setting (0.00595). This ordering contrasts with the ILI experiments, where large, high-capacity foundation-style architectures often dominate: here, with only three years of data, a smaller recurrent model (TinyLSTM) provides the strongest pure-hospitalization baseline.

Adding ILI as an auxiliary covariate improves performance for nearly all architectures. For TimeLLM, ILI input reduces the average MSE from 0.00595 to 0.00433, indicating that even a token-based model can leverage ILI as a helpful side signal. The effect is more consistent and pronounced for numeric models. TFT improves from 0.00320 to 0.00295 on average, LSTM from 0.00329 to 0.00269, and TinyLSTM from 0.00262 to 0.00237. These gains are especially visible at the longer horizons: for example, TinyLSTM’s week-4 MSE drops from 0.00611 to 0.00560 when ILI is added, and LSTM’s from 0.00641 to 0.00589. PatchTST and iTransformer also benefit, though more modestly (average MSE improving from 0.00361 to 0.00343 and from 0.00345 to 0.00344, respectively). Overall, these results support the view that ILI acts as a complementary indicator for hospitalization, helping models resolve the phase and intensity of recent waves, particularly at 3–4-week horizons where hospitalization dynamics are most uncertain.

MultiFoundationCore, applied here with ILI as an input stream, achieves the best overall performance on this task. Its average MSE of 0.00224 is lower than any single model, including the strong TinyLSTM+ILI baseline (0.00237). The improvements at weeks 3 and 4 are modest but consistent: MFC attains week-3 and week-4 MSEs of 0.00333 and 0.00523, compared to 0.00355 and 0.00560 for TinyLSTM+ILI. Given the limited length of the hospitalization dataset, these gains demonstrate that ensembling pretrained forecasters and fusing them with ILI covariates can still deliver a measurable advantage, even when simpler recurrent models already perform well.

This experiment highlights three points. First, short hospitalization histories favor relatively compact architectures (such as TinyLSTM) over very large foundation-style models. Second, ILI provides a meaningful additional signal for hospitalization forecasting: augmenting models with ILI consistently reduces multi-week errors, especially at longer horizons. Third, MultiFoundationCore is able to exploit ILI most effectively, yielding the lowest overall errors despite the limited data regime. Combined with our earlier results—where hospitalization helped improve ILI prediction—this bidirectional relationship underscores the value of jointly modeling multiple, mechanistically linked influenza indicators.

\subsection{Scope and Limitations}

This study compares the point-forecasting behavior of modern time series models for regional influenza forecasting using ILI and hospitalization data under CDC-aligned 1--4-week horizons. Our goal is to understand how different model families, pretraining sources, auxiliary signals, and generalization settings affect forecasting performance, rather than to build a complete real-time forecasting system. Therefore, we focus on deterministic point forecasts and evaluate models using MSE and NNSE. The spatial split should also be interpreted as a complementary transfer test, since the evaluation is limited to the available HHS regions. Probabilistic forecasting, uncertainty quantification, calibration analysis, and broader spatial generalization are important next steps, and we discuss these directions in the following section.

\section{Conclusion}
\label{conclusion}

Seasonal influenza forecasting sits at the intersection of public health, scientific modeling, and machine learning practice. In this work, we presented what is, to our knowledge, the first systematic study of modern time series foundation models and large pattern architectures for influenza forecasting. Using regional ILI and influenza hospitalization datasets, standardized preprocessing, and both temporal (within-region) and spatial (across-region) generalization regimes, we compared 17 deep time series models under realistic 1–4-week horizons that mirror operational FluSight practice. This setting exposed strengths and weaknesses that are largely invisible in the stylized, long-horizon, single-series ILI benchmarks that dominate the current foundation-model literature.

Building on this benchmark, we show our MoE-style foundation-model framework, \emph{MultiFoundationCore}, consistently achieves best performance on 1–4-week regional ILI forecasts in the temporal setting, outperforming all individual foundation and pattern baselines in both MSE and NNSE. When applied to normalized influenza hospitalization prediction—a shorter, noisier series where compact recurrent models perform surprisingly well—MultiFoundationCore still provides measurable gains, particularly at multi-week horizons. 

Our experiments show that numerically oriented time series foundation models such as PatchTST and iTransformer are highly effective for ILI forecasting, especially when pre-trained on large or related corpora such as M4, TrafficL, and hospitalization series. Cross-domain pre-training systematically improves medium-range forecasts, with the largest gains at the more challenging 3–4-week horizons. In contrast, LLM-style approaches such as TimeLLM lag behind well-tuned numerical architectures in this regime, even after fine-tuning. This underperformance should not be interpreted only as a model-specific failure; rather, it indicates a broader mismatch between language-oriented sequence modeling and continuous epidemic forecasting, where numerical scale, temporal phase, local slope, seasonality, and mechanistically related signals carry essential predictive information.

Our analyses also clarify how data availability and auxiliary signals shape performance. Longer ILI archives substantially improve medium-range accuracy, underscoring the value of preserving and curating multi-decade surveillance records. Influenza hospitalizations help in both directions: as a pre-training domain, they strengthen ILI forecasts; as an auxiliary covariate, they improve hospitalization prediction when combined with ILI. We further show that moderate retraining frequencies strike a practical balance between computational cost and performance, with more frequent temporal adaptation especially beneficial at 3–4-week horizons.

Looking forward, our work opens several avenues for future research. The main goal of this paper was to understand the key forecasting behavior of time series foundation models under realistic influenza forecasting settings, rather than to develop a fully operational real-time forecasting system. A natural next step is to extend MultiFoundationCore and our benchmark to probabilistic forecasting, including uncertainty quantification and calibration metrics, which would bring the study even closer to the FluSight and Forecast Hub ecosystems. Second, integrating mechanistic or causal structure—such as SEIR-style components, mobility, interventions, and demographic features—into foundation-style backbones may further improve robustness under regime shifts and pandemics. Third, expanding beyond influenza to other pathogens, countries, and surveillance systems will test the portability of our conclusions and the generality of MultiFoundationCore as a time series foundation model for public health. Finally, our released code, splits, and pretrained checkpoints provide a reusable testbed for the community to explore new architectures, training strategies, and data sources in a controlled, CDC-aligned setting. We hope this work helps bridge the gap between generic time series foundations and the specific, high-stakes demands of real-time epidemic forecasting.

\section*{Acknowledgement}
This work was supported in part by the National Science Foundation under Grant 2346173, Grant 2411009, and Grant CCF-1918656; in part by the CSTE--CDC under Award PO8968; in part by the Defense Threat Reduction Agency under Award HDTRA1-24-R-0028; in part by the Centers for Disease Control and Prevention (CDC) and DCLS under Cooperative Agreement 6NU50CK000555-03-01; and in part by the MIDAS Coordination Center under Award R24GM1532920.

\bibliographystyle{IEEEtran}
\bibliography{refs}

% \FloatBarrier

\appendix

% --- top part: fake two-column layout, appendix text on right ---
\noindent
\begin{minipage}[t]{0.48\textwidth}
\vspace{0pt}
% empty left side
\end{minipage}
\hfill
\begin{minipage}[t]{0.48\textwidth}
\vspace{0pt}
\section*{Models' Configurations and Hyperparameters}

This appendix reports the implementation details and hyperparameter settings used to reproduce the main forecasting experiments. Table~\ref{tab:appendix_table_i_model_configs} summarizes the model configurations for the primary temporal ILI comparison in Table~I, including the implementation package, forecasting strategy, prediction horizon, input window length, maximum training steps, and learning rate. These entries distinguish direct multi-output models from iterative forecasters, indicate which models were evaluated from pretrained checkpoints rather than trained from scratch, and document model-specific choices such as the shorter expert-stream input window used by MultiFoundationCore. Table~\ref{tab:appendix_architecture_defaults_all_models} complements this information by listing the retained architecture defaults for the principal forecasting models, including NeuralForecast-based models, custom TensorFlow/Keras baselines, Chronos checkpoints, and the Statsmodels ARIMA baseline.
\end{minipage}

% --- full-width table, non-floating ---
\begin{table*}[!b]
\centering
\caption{Training and evaluation configurations for models used in Table I.}
\label{tab:appendix_table_i_model_configs}
\begin{tabular}{lcccccc}
\hline
\textbf{Model} &
\textbf{Implementation} &
\textbf{Strategy} &
\textbf{Horizon} &
\textbf{Input size} &
\textbf{Training steps} &
\textbf{Learning rate}  \\
\hline
ARIMA & Statsmodels & Iterative & 4 & 104 & -- & -- \\
TCN & NeuralForecast & Direct & 4 & 52 & 400 & $1\times10^{-3}$ \\
TiDE & NeuralForecast & Direct & 4 & 52 & 3000 & $1\times10^{-3}$ \\
VanillaTransformer & NeuralForecast & Direct & 4 & 52 & 500 & $1\times10^{-3}$ \\
TimeLLM & NeuralForecast & Direct & 4 & 52 & 400 & $1\times10^{-3}$ \\
TimesNet & NeuralForecast & Direct & 4 & 52 & 200 & $1\times10^{-3}$ \\
TFT & NeuralForecast & Direct & 4 & 52 & 200 & $1\times10^{-3}$ \\
iTransformer & NeuralForecast & Direct & 4 & 52 & 75 & $1\times10^{-4}$ \\
PatchTST & NeuralForecast & Direct & 4 & 52 & 150 & $5\times10^{-4}$ \\
Chronos-T5  & Pretrained checkpoint & Iterative & 4 & 52 & -- & -- \\
Chronos-Bolt  & Pretrained checkpoint & Iterative & 4 & 52 & -- & -- \\
LSTM-direct & Custom TensorFlow/Keras & Direct & 4 & 52 & 100 & $1\times10^{-4}$ \\
LSTM-iterative & Custom TensorFlow/Keras & Iterative & 4 & 52 & 100 & $1\times10^{-4}$ \\
MultiFoundationCore & Custom TensorFlow/Keras & Direct & 4 & 10 & 175 & $1\times10^{-4}$ \\
\hline
\end{tabular}

\vspace{0.5em}
\begin{flushleft}
\footnotesize
All trainable models used MSE loss. For NeuralForecast models, the input size denotes the autoregressive look-back window in weeks. ARIMA was implemented with the Statsmodels and fit independently at each rolling forecast origin using a 104-week history; because it is a classical statistical model, training steps and learning rate are not applicable. Chronos models were evaluated from pretrained checkpoints rather than trained from scratch. The LSTM-direct baseline was trained to produce 1--4-week forecasts directly, whereas the LSTM-iterative baseline was trained for one-step prediction and recursively rolled forward to produce 1--4-week forecasts. MultiFoundationCore uses a shorter input window because its inputs are forecast-derived expert streams rather than only raw weekly ILI values. Unless otherwise specified, the reported performance values in Table~I are averaged over at least five independent runs. For detailed information and additional experimental configurations, please refer to our GitHub repository \textit{\url{https://github.com/alireza-jafari/Epidemic-Times-Series-Foundation-Models-Benchmark}}.
\end{flushleft}
\end{table*}

\begin{table*}[t]
\centering
\footnotesize
\caption{Architecture and configuration values retained for the main forecasting models used in the paper. NeuralForecast models were implemented using the NeuralForecast package~\cite{olivares2022neuralforecast} and ARIMA was implemented using the Statsmodels~\cite{seabold2010statsmodels}. We manually specified experiment-level settings such as horizon, input size, learning rate, maximum training steps, batch size, and random seed.}
\label{tab:appendix_architecture_defaults_all_models}
\renewcommand{\arraystretch}{1.15}
\begin{tabularx}{\textwidth}{p{0.19\textwidth}p{0.16\textwidth}>{\raggedright\arraybackslash}X}
\hline
\textbf{Model} & \textbf{Implementation} & \textbf{Default architecture or configuration values retained} \\
\hline

PatchTST~\cite{nie2023time} &
NeuralForecast &
Transformer encoder with patching and channel independence. Defaults retained: encoder layers = 3; attention heads = 16; hidden size = 128; linear hidden size = 256; patch length = 16; stride = 8; dropout = 0.2; fully connected dropout = 0.2; head dropout = 0.0; attention dropout = 0.0; RevIN enabled; RevIN affine disabled; subtract-last RevIN enabled; residual attention enabled; learnable positional embedding enabled; GELU activation. \\

TFT~\cite{lim2021temporal} &
NeuralForecast &
Temporal Fusion Transformer with recurrent local processing and attention-based temporal fusion. Defaults retained: hidden size = 128; attention heads = 4; attention dropout = 0.0; gated residual network activation = ELU; recurrent encoder type = LSTM; recurrent layers = 1; shared initial recurrent state disabled; dropout = 0.1; target size = 1. Static and future exogenous variables were not used; hospitalization was used only as a historical exogenous input in the input-stream variant. \\

TimeLLM~\cite{jin2024timellmtimeseriesforecasting} &
NeuralForecast &
LLM-reprogramming forecaster using GPT-2 as the default language-model backbone. Defaults retained: patch length = 16; stride = 8; feed-forward dimension = 128; top-$k$ lag tokens = 5; LLM hidden dimension = 768; model dimension = 32; attention heads = 8; encoder input size = 7; decoder input size = 7; LLM hidden layers = 32; LLM output attention enabled; LLM output hidden states enabled. \\

iTransformer~\cite{liu2023itransformer} &
NeuralForecast &
Inverted Transformer architecture that treats variates as tokens. Defaults retained: hidden size = 512; attention heads = 8; encoder layers = 2; decoder layers = 1; feed-forward dimension = 2048; attention factor = 1; dropout = 0.1; normalization enabled. \\

TiDE~\cite{das2023long} &
NeuralForecast &
Dense encoder--decoder forecasting model based on MLP residual blocks. Defaults retained: hidden size = 512; decoder output dimension = 32; temporal decoder dimension = 128; dropout = 0.3; layer normalization enabled; encoder layers = 1; decoder layers = 1; temporal width = 4. \\

TCN~\cite{bai2018empiricalevaluationgenericconvolutional} &
NeuralForecast &
Temporal Convolutional Network with dilated convolutional encoder and MLP decoder. Defaults retained: kernel size = 2; dilations = [1, 2, 4, 8, 16]; encoder hidden size = 128; encoder activation = ReLU; context size = 10; decoder hidden size = 128; decoder layers = 2. \\

LSTM~\cite{hochreiter1997long} &
Custom  &
Encoder--decoder LSTM with MLP decoder. Defaults retained: training horizon length $h_{\mathrm{train}}=1$; encoder layers = 2; encoder hidden size = 128; encoder bias enabled; encoder dropout = 0.0; decoder hidden size = 128; decoder layers = 2; direct multi-output forecasting used when \texttt{recurrent=False}. For the iterative LSTM variant, forecasts were generated recursively. \\

ARIMA~\cite{box2015time} &
statsmodels &
Classical autoregressive integrated moving-average baseline implemented with \texttt{statsmodels.tsa.arima.model.ARIMA}. The model was fit separately at each rolling forecast origin and generated recursive 1--4-week-ahead forecasts. The configuration used ARIMA$(1,0,0)$. A constant trend term was enabled, stationarity and invertibility constraints were relaxed, and a persistence forecast was used only as a fallback when fitting failed or produced invalid forecasts. \\

TimesNet~\cite{wu2022timesnet} &
NeuralForecast &
2D temporal-variation model using FFT-based period discovery and Inception-style convolutional blocks. Defaults retained: hidden size = 64; convolutional hidden size = 64; dropout = 0.1; top-$k$ periods = 5; number of convolution kernels = 6; encoder layers = 2. \\

VanillaTransformer~\cite{vaswani2017attention,zhou2021informer} &
NeuralForecast &
Full-attention Transformer baseline following the Informer-style implementation. Defaults retained: decoder input-size multiplier = 0.5; hidden size = 128; dropout = 0.05; attention heads = 4; convolutional hidden size = 32; activation = GELU; encoder layers = 2; decoder layers = 1; full attention used in encoder and decoder. \\

TSMixer~\cite{chen2023tsmixerallmlparchitecturetime} &
NeuralForecast &
MLP-based multivariate model with repeated temporal and feature mixing. Defaults retained: number of mixing blocks = 2; feed-forward dimension = 64; dropout = 0.9; RevIN enabled. This model was used as one of the expert streams in the MultiFoundationCore setting. \\

Chronos~\cite{ansari2024chronos} &
Pretrained checkpoint &
Used as pretrained zero-shot or lightly adapted foundation models rather than models trained from scratch with user-defined architecture hyperparameters. Chronos-Bolt is based on a T5 encoder--decoder architecture and uses patching plus direct multi-step decoding; original Chronos-style models use language-model tokenization of time-series values. Model size was selected by checkpoint name, e.g., mini, small, base, or large. \\

TinyLSTM &
Custom  &
Compact recurrent baseline used for the short-history hospitalization experiment. Unlike the NeuralForecast defaults above, TinyLSTM is a custom small-capacity model designed to reduce overfitting in the three-year hospitalization setting. \\

MultiFoundationCore~\cite{jafari2024time, Junyang2026} &
Custom  &
Cross-attention mixture-of-experts fusion model. In Table VI, the model fused expert forecasts from PatchTST, TSMixer, TFT, iTransformer, TimeLLM, and VanillaTransformer. Each expert contributed 1--4-week-ahead forecasts, giving $6 \times 4 = 24$ forecast-derived input channels. The model also used an ILI history channel and, in the input-stream variant, an additional hospitalization channel. \\

\hline
\end{tabularx}
\end{table*}

% \begin{table*}[t]
% \centering
% \caption{MultiFoundationCore architecture and training configuration for Table VI.}
% \label{tab:appendix_mfc_hyperparams}
% \begin{tabular}{ll}
% \hline
% \textbf{Component} & \textbf{Configuration} \\
% \hline
% Fusion model type & Cross-attention mixture-of-experts model \\
% Expert models & PatchTST, TSMixer, TFT, iTransformer, VanillaTransformer, TimeLLM \\
% Number of expert models & 6 \\
% Forecast horizons per expert & 4 \\
% Forecast-derived channels & $6 \times 4 = 24$ \\
% Historical target channel & ILI look-back channel \\
% Auxiliary channel & Hospitalization, used only in the input-stream variant \\
% Input channels without hospitalization & 25 \\
% Input channels with hospitalization & 26 \\
% Input window length & 10 time steps \\
% Output dimension & 4 horizons \\
% Per-channel encoder & LSTM with 8 hidden units \\
% Target-history encoder & LSTM with 8 hidden units \\
% Attention layer & Multi-head attention \\
% Attention heads & 8 \\
% Key dimension & 16 \\
% Value dimension & 16 \\
% Post-attention layer & Dense layer with 16 ReLU units \\
% Output layer & Dense layer with 4 linear outputs \\
% Optimizer & Adam \\
% Learning rate & $2\times 10^{-4}$ \\
% Loss & MSE \\
% Batch size & 32 \\
% Epochs & 700 \\
% Random seed & 100 \\
% Training samples & 1100 \\
% Test samples & 260 \\
% \hline
% \end{tabular}
% \end{table*}

\end{document}